\documentclass[lettersize,journal]{IEEEtran}
\usepackage{amsmath,amsfonts}

\newcommand{\normSI}[1]{\lvert #1 \rvert}
\newcommand{\normDB}[1]{\lvert \lvert #1 \rvert \rvert}
\usepackage{algorithmic}
\usepackage{algorithm}
\usepackage{array}
\usepackage[caption=false,font=normalsize,labelfont=sf,textfont=sf]{subfig}
\usepackage{textcomp}
\usepackage{stfloats}
\usepackage{url}
\usepackage{verbatim}
\usepackage{graphicx}
\usepackage{booktabs}
\usepackage{multirow}
\usepackage{float}
\usepackage{cite}
\usepackage{censor}
\usepackage{etoolbox}
\usepackage{lipsum}
\usepackage[table,xcdraw]{xcolor}
\begin{document}

\title{Flight through Narrow Gaps with Morphing-Wing Drones}

\author{Julius Wanner$^{1}$, Hoang-Vu Phan$^{2}$, Charbel Toumieh$^{1}$, Dario Floreano$^{1}$%
\thanks{$^{1}$ J. Wanner, C. Toumieh and D. Floreano are with the Laboratory of Intelligent Systems, École Polytechnique Fédérale de Lausanne (EPFL), Lausanne, Switzerland {\tt\small (julius.wanner@epfl.ch, charbel.toumieh@epfl.ch, dario.floreano@epfl.ch)}}
\thanks{$^{2}$ H.V. Phan is with the Avian-Insect Robotics in Motion Laboratory, University of Nevada, Reno, USA {\tt\small (vuphan@unr.edu)}}%
}

% \thanks{The authors are with
% the Laboratory of Intelligent Systems, Ecole Polytechnique Federale de Lausanne (EPFL), CH-1015 Lausanne, Switzerland (E-mail:
% julius.wanner@epfl.ch, vu.phan@epfl.ch, charbel.toumieh@epfl.ch, dario.floreano@epfl.ch).}}% <-this % stops a space
% \thanks{Manuscript received April 19, 2021; revised August 16, 2021.}}

% The paper headers
%\markboth{IEEE Transactions on Robotics (T-RO)}%
% \markboth{Journal of \LaTeX\ Class Files,~Vol.~14, No.~8, August~2021}%
% {Shell \MakeLowercase{\textit{J. Wanner et al.}}: Flight through Narrow Gaps with Morphing-Wing Drones}

% \IEEEpubid{0000--0000/00\$00.00~\copyright~2021 IEEE}
% Remember, if you use this you must call \IEEEpubidadjcol in the second
% column for its text to clear the IEEEpubid mark.

\makeatletter
\g@addto@macro\@maketitle{\vspace{-2em}}
\makeatother
\maketitle

\makeatother

\begin{abstract}
 
The size of a narrow gap traversable by a fixed-wing drone is limited by the drone's wingspan. Inspired by birds, here, we enable the traversal of a gap of sub-wingspan width and height using a morphing-wing drone capable of temporarily sweeping in its wings mid-flight. This maneuver poses control challenges due to sudden lift loss during gap-passage at low flight speeds and the need for precisely timed wing-sweep actuation ahead of the gap. To address these challenges, we first develop an aerodynamic model for general wing-sweep morphing drone flight including low flight speeds and post-stall angles of attack. We integrate longitudinal drone dynamics into an optimal reference trajectory generation and Nonlinear Model Predictive Control framework with runtime adaptive costs and constraints. Validated on a 130 g wing-sweep-morphing drone, our method achieves an average altitude error of 5 cm during narrow-gap passage at forward speeds between 5–7 m/s, whilst enforcing fully swept wings near the gap across variable threshold distances. Trajectory analysis shows that the drone can compensate for lift loss during gap-passage by accelerating and pitching upwards  ahead of the gap to an extent that differs between reference trajectory optimization objectives. We show that our strategy allows for accurate gap passage on hardware whilst maintaining a constant forward flight speed reference and near-constant altitude.

% \lipsum[1-2]
\end{abstract}


\section{Introduction}

\IEEEPARstart{B}{irds} are capable of performing agile maneuvers by morphing their wings in flight \cite{ros_rules_2017, badger_sideways_2023}. Therefore, there is increasing interest in translating mechanisms of bird wing morphing to advance the design of similar-scale drones \cite{harvey_review_2022}. 
%To replicate flight strategies leveraged by birds, studies on a class of morphing drones have emerged. 
For example, benefits of wing-sweep morphing in roll control and drag modulation were highlighted before \cite{di_luca_bioinspired_2017}, while other designs \cite{ajanic_bioinspired_2020, ajanic_sharp_2022, phan_twist_2024, chang_soft_2020} were developed to study how individual wing- and tail- morphing actuation strategies can enhance the flight envelope either in cruise or aggressive turn scenarios. Additional studies controlled morphing actuators to perform perching maneuvers similar to birds \cite{wuest_agile_2024} and to loiter at high angles of attack\cite{zhang_enhancing_2024}. Avian flight researchers have furthermore shown that birds tuck their wings to fly through narrow gaps and adapt their flight path ahead of the gap to compensate for the altitude loss caused by the reduction of the aerial surface. \cite{schiffner_minding_2014, vo_anticipatory_2016, altshuler_comparison_2018}. 
However, a translation of this strategy to morphing-winged drones to allow for flight through narrow gaps remains unexplored.

\begin{figure}[t]
    \centering    \includegraphics[width=\columnwidth]{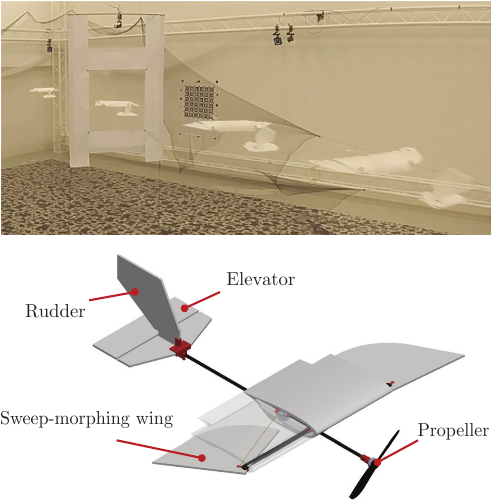}
    \caption{A wing-sweep morphing drone flies through a narrow gap by temporarily sweeping in its wings.}
    \label{fig:title_image}
\end{figure}
\vspace{10pt}

Controlled flight through narrow gaps has previously been achieved with foldable rotary-wing drones, which are capable of hovering and flying at low speed  \cite{falanga_aggressive_2017, falanga_foldable_2019, bucki_design_2019, riviere_agile_2018}. Because the thrusters point upward, the quadrotor's lift and thrust remain unchanged while the shape is changed. This lets the drone pass through gaps without affecting its upwards or forward acceleration during closed-loop flight. In contrast, attempts of flight with fixed-wing-drones through gaps narrower than their wingspan have required alternative approaches. A "knife-edge maneuver", where the drone momentarily reaches a roll angle of more than 70° when passing through a tight gap, was previously adopted  \cite{barry_flying_2014}. However, during this maneuver, the wings generate less lift and cause an altitude loss that can be difficult to recover. Furthermore, this maneuver is feasible only if the gap offers sufficient vertical clearance for the wings to pass. Other researchers developed a drone with passively folding wings that collapse when hitting the gap edges \cite{henry_wing_2023}, but this strategy can produce unrecoverable destabilization if the impact is stronger or earlier on one side than on the other.

Here, we study an avian-inspired strategy that consists of temporarily reducing the wingspan by sweeping the wing tips inwards while passing through narrow gaps. This morphing strategy does not require high vertical clearance through the gap and reduces the risk of collisions with the gap edges. However, it requires rapid wing-sweep actuation and control strategies that can preemptively compensate for altitude loss resulting from fully sweeping in the wings. This is particularly relevant at low 
speeds and operating conditions near or beyond stall. 

We validate this approach on a new morphing-wing drone with a lightweight wing-sweep mechanism. Differently from birds, the drone is powered by a frontal propeller that can produce forward thrust through the gap and has a conventional aircraft tail. We develop a parametric model capturing aerodynamic loads for variable wing sweep configuration across low-Reynolds flight regimes at both pre- and post-stall angles of attack. Using a longitudinal drone dynamics model, we generate optimal gap passage trajectories under optimization criteria of minimizing the altitude change, speed change, and the time during which the wings remain fully swept. We track these trajectories with a Nonlinear Model Predictive Control (MPC) strategy that leverages stage-dependent actuator constraints and costs to enforce fully swept wings on hardware in real-time. Lateral alignment is controlled separately by a cascaded controller with an adaptive control allocation matrix.

Our results show that the trajectory optimization objective primarily governs variations in flight speed and pitch angle at the gap, while altitude remains near constant. When speed changes are minimized, the drone relies on pitch adjustments to traverse the gap without significant altitude variation. Closed-loop experiments confirm collision-free passage with centimeter-level accuracy at 5–7 m/s, with inward wing sweeping initiated 0.2–0.6 m ahead of the gap. These findings establish a basis for extendable model-based control architectures applied to wing-sweep morphing drones that must negotiate narrow gaps in cluttered environments.

% \vspace{10pt}

\section{Morphing-wing drone design} % (1 page)
% \subsection{Drone design} %(0.5 - 1 page)

Avian-inspired morphing-wing drones often use overlapping artificial or natural feathers \cite{ajanic_sharp_2022, ajanic_bioinspired_2020, di_luca_bioinspired_2017, phan_twist_2024, chang_soft_2020}. These fold in and out to accommodate wing sweep changes but require mechanically complex and heavier structures. The added outboard wing mass demands more powerful actuators, increasing system weight further, and hindering low-speed maneuverability. Reducing the mass of the wing-sweeping mechanism and the outboard wing allows for a lighter system and faster wing-sweep actuation, improving maneuverability at slow speeds and reducing actuator response times for flight in cluttered environments with narrow-gaps.

\begin{figure}[ht]
    \includegraphics[width=\columnwidth]{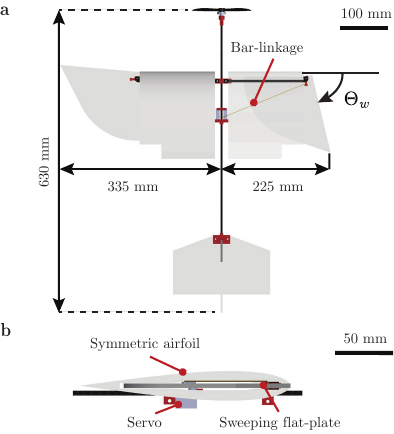}
    \caption{The 130 g morphing-wing drone platform leveraging wing-sweep morphing for narrow gap passage. \textbf{(a)} The outboard wing sweep is actuated by a servo-driven bar-linkage mechanism. The sweep angle $\Theta_w$ varies between -5° and 75° from the fully extended to the fully swept configuration respectively. \textbf{(b)} The side profile view shows the outboard flat-plate sweeping into the symmetric root airfoil.}
    \label{fig:drone_description}
\end{figure}

We hence develop a new 130-gram ready-to-fly wing-sweep drone [Fig.  \ref{fig:title_image}]. Each wing-half is made of expanded polypropylene (EPP) and consists of two parts: a root wing, which is attached to the fuselage and has a symmetric S8035 airfoil profile, and an outer wing segment with a flat-plate profile that can sweep backward inside the root wing [Fig.  \ref{fig:drone_description}]. The outer wing segment has a mass of approximately 4 grams and is actuated by a 4.5 g Bluebird BMS-101DMG servo motor through a bar linkage mechanism [Fig. \ref{fig:drone_description}]. The outer wing segment can sweep over a range of 80° from fully-extended to fully swept configuration [Fig. \ref{fig:drone_description}] in 0.10 s. One half-wing comprising the root wing, the outboard wing and the sweep actuation mechanism, excluding the servo, has a mass of 25 g. The left and right outer wing segments can be swept independently by separate servos. Hence, through symmetric actuation of the two servos, the total wing area can be changed, while through asymmetric actuation, a rolling motion can be induced. The wingspan is 0.67 meters in the extended configuration and 0.45 meters in the fully swept, corresponding to a wingspan reduction of 34$\%$. For pitch and yaw control, we use a conventional tail with an elevator and a rudder that are independently actuated by servos positioned on the tail. The elevator surface ratio is $\sim 60 \%$ of the total horizontal tail area to guarantee high pitch-rate control authority at low flight speeds. Forward thrust is generated by a 9-gram Pichler Nano 9G brushless motor, which provides up to 1.2 N of thrust with a 6 x 3 inch propeller when powered by a 200 mAh 2S lithium-polymer battery. The motor is controlled by a Pichler XQ6 LT electronic speed controller (ESC) with integrated auxiliary 5V power output. An ESP32 microcontroller is used to receive wireless (WiFi) servo input commands from an off-board controller. The microcontroller handles the transmission of Pulse Width Modulation (PWM) signals to the four servos (left wing sweep, right wing sweep, rudder, elevator) and the ESC.

\section{Wing-sweep morphing drone dynamics}
\subsection{Equations of motion}
\label{sec:dynamics}

To generate longitudinal optimal reference trajectories for the narrow-gap passage problem and enable accurate tracking with model-based controllers, we develop a nonlinear state-transition model of the wing-sweep morphing drone. We describe the longitudinal equations of motion (\ref{eqn:dynamics_eom}) with a state vector $\mathbf{x}$ and an input vector $\mathbf{u}$. The inputs contain a normalized motor input $u_m$, the elevator change rate $u_e$, and a normalized wing-sweep input $u_{w}$. The longitudinal and vertical velocity states are defined in body-frame, denoted respectively within the vector $\mathbf{v}_\mathit{b} = [u,w]^\top$. The pitch rate is described by $q$. The spatial states contain the forward and vertical positions and the pitch angle in the inertial frame, defined as $x$, $z$, and $\theta$, respectively. The thrust dynamics are described using a time-varying motor speed state $\Omega_{m}$, updated through a first-order model and a polynomial dependency on $u_m$. Furthermore, the normalized elevator angle state $x_{e}$ is described by a first-order linear model of the form $\dot{x}_{e} = u_e$. By choosing to directly control the elevator angle rate input, we can enforce rate limits of the elevator angle change and thereby reduce destabilizing oscillations in subsequent closed-loop control. The normalized sweep states are described by an identified second-order linear model with state vector $\mathbf{x}_w = [x_{w}, \dot{x}_{w}]^\top$. 

\begin{align}
\label{eqn:dynamics_eom}
    & \mathbf{x} = [x, z, \theta, u, w, q, \Omega_m, x_{w}, \dot{x}_{w}, x_{e}]^\top \nonumber \\
    & \mathbf{u} = [u_m, u_e, u_{w}]^\top \nonumber \\
    & \mathbf{\dot{v}}_b = \frac{1}{m}(\mathbf{f}_{w,\mathit{b}} + \mathbf{f}_{t,\mathit{b}} + \mathbf{f}_{m,\mathit{b}}) + {\mathbf{R(\theta)}}\mathbf{g}_0 - \mathbf{v}_{b,\text{cor}} \\
    & \dot{q} = \frac{1}{\overline{I_{yy}}}(\tau_{w,\mathit{b}} + \tau_{t,\mathit{b}} + \tau_{m,\mathit{b}}) \nonumber \\
    & \mathbf{r}_\text{cg} = [x_\text{cg}(\Theta_{w}), z_\text{cg}] \nonumber \\
\nonumber
\end{align}

The aerodynamic forces acting on both wing and tail are given by $\mathbf{f}_{w,\mathit{b}}$ and $\mathbf{f}_{t,\mathit{b}}$ respectively and the corresponding induced pitch moments are denoted as $\tau_{w,\mathit{b}}$ and $\tau_{t,\mathit{b}}$. The additive Coriolis term due to derivation in the body frame is given as $\mathbf{v}_{b,\text{cor}} = [qw, -qu]^\top$. The thrust force vector in drone body frame is given as $\mathbf{f}_{m,\mathit{b}} = [T_m, 0]^\top$, acting in the drone longitudinal axis with a magnitude $T_m$. Induced pitch moments by an offset between the thruster and center of gravity are captured by $\tau_{m,b}$. The gravitational force vector in inertial frame is given as $\mathbf{g}_0$ and rotated into body frame with the two-dimensional rotation matrix $\mathbf{R(\theta)} \in \mathbb{R}^{2\times 2}$. Due to the low mass of the outboard wings, the longitudinal mass moment of inertia varies by $< 0.5 \%$ between the symmetrically extended and symmetrically fully swept configuration. Consequently, for simplicity, $\overline{I_{yy}}$ is an average longitudinal mass moment of inertia taken at the mid wing sweep angle location and assumed as constant for this drone design. As the outboard wing mass is furthermore low, we neglect dynamic moments or accelerations caused by actuator momentum effects. The planar center of gravity is denoted as $\mathbf{r}_\text{cg}$, with a constant vertical location at $z_\text{cg}$ relative to the leading-edge of the wing. The longitudinal center of gravity location $x_\text{cg}(\Theta_w)$ relative to the leading edge of the wing is given as a linear function of the wing sweep angle and only varies by up to $\pm$ 2 mm due to wing-sweep morphing. 

\subsection{Aerodynamic loads}
\label{sec:aero_model}
High-fidelity characterization of aerodynamic loads on morphing-wing drones has been pursued through Computational Fluid Dynamics (CFD) simulations \cite{li_review_2018, obradovic_modeling_2011}. While accurate, CFD is computationally expensive and unsuitable for use in real-time, model-based control. As a result, empirical wind-tunnel testing was previously used to directly characterize aerodynamic coefficients for a physical prototype across static flight conditions and multiple morphologies \cite{wuest_agile_2024, liu_employing_2023}. However, such campaigns are time-consuming and specific to a single design, requiring repetition for each new geometry, wind speed, angle of attack, and actuator configuration to yield a model at discrete points of operation. To address these limitations, we present and validate a real-time computable aerodynamic model parameterized by wing sweep angle, wing geometry, and a minimal set of empirically or numerically identifiable airfoil properties. This model allows for adaptability to different morphing-drone geometries and accounts for thrust slipstream and low-Reynolds number lift degradation in both pre- and post-stall regimes.

To estimate the aerodynamic loads that act on the wings, we begin by describing the pre-stall aerodynamics of a morphing wing. When the wings sweep symmetrically by an angle $\Theta_w$ [see Fig. \ref{fig:drone_description}], the wingspan $b_{w}(\Theta_w) $ and surface area $S_{w}(\Theta_w) $ are impacted, yielding a variable aspect ratio $\Lambda_w(\Theta_w) = {b_{w}}^2/S_{w}$. The aspect ratio $\Lambda_w(\Theta_w)$ ranges from 3-5 for our drone. Given an airfoil profile with a known pre-stall lift curve slope $c_{l_\alpha}$, the lift and drag coefficients $c_{L,\text{pre}}$ and $c_{D,\text{pre}}$ at an angle of attack $\alpha$ in the pre-stall regime ($\alpha \leq \alpha_\text{st}$) for wings of small drones are obtained assuming approximate elliptical-wing behavior \cite{beard_small_2012}.
To obtain the coefficients $c_{L,\text{st}}$ and $c_{D,\text{st}}$ in the post-stall regime ($\alpha > \alpha_\text{st}$), we assume two-dimensional flat-plate airfoil aerodynamics \cite{cory_experiments_2008} and apply a correction for effects on a finite three-dimensional wing \cite{selig_real-time_2014}. To account for lift coefficient degradation at Reynolds numbers below $\text{Re}_\text{nom} = 1 \times 10^5$ \cite{winslow_basic_2018}, we introduce a degradation factor $f_\text{Re}$ described by a power law. The resulting formulations for lift and drag coefficients are formally described in (\ref{eq:aero_coefficients}).

\begin{align}
\label{eq:aero_coefficients}
    c_{L,\text{pre}} &= f_\text{Re}(\text{Re}_{w}) \left[c_{l,0} + c_{l_\alpha} \frac{\Lambda_w(\Theta_w)}{2 + \sqrt{4 + {\Lambda_{w}(\Theta_w)}^2}}\alpha \right] \nonumber \\
    c_{D,\text{pre}} &= c_{D,0} + \frac{c_{L,\text{pre}}^2}{\pi \Lambda_w(\Theta_w)} \nonumber \\
    c_{L,\text{st}} &= 2{f_\text{Re}(\text{Re}_{w})}\sin{\alpha}\cos{\alpha}[1 - K(\alpha) (1-k_{C_d})] \nonumber \\
    c_{D,\text{st}} &= 2{f_\text{Re}(\text{Re}_{w})}\sin^2{\alpha}[1 - K(\alpha) (1-k_{C_d})] \nonumber \\
    % \\
    % \text{where:}
    % \\
    K(\alpha) &= \cos(\pi\frac{\normSI{\alpha} - \alpha_\text{st}}{\pi - \alpha_\text{st}} - \frac{\pi}{2}) \\
    k_{C_d}(\Theta_w) &= 1 - 0.41(1 - e^{-17 / \Lambda_w (\Theta_w)}) \nonumber \\
    {f_\text{Re}}(\text{Re}_{w}) &= 1 - \left[1 - \min{\left(\frac{\text{Re}_w}{\text{Re}_{\text{nom}}}, 1\right)} \right]^a \nonumber \\
    \text{Re}_{w}(V,\Theta_w) &= \frac{\rho V \overline{c}(\Theta_w)}{\mu} \nonumber \\
\nonumber
\end{align}

For the presented prototype, we assume $c_{l,0} = 0$ due to its symmetric wing, with $c_{l_\alpha} = 2\pi$ for fully attached flow conditions according to thin airfoil-theory. The parasitic drag coefficient $c_{D,0}$ is obtained from measurement at a zero angle-of-attack and the lift-degradation coefficient $a$ empirically from lift and drag measurements at three wind speeds and a constant wing sweep configuration. The Reynolds number across the wing is denoted as $\text{Re}_{w}$, considering the mean aerodynamic chord $\overline{c}$ dependent on the symmetric wing sweep angle, the averaged inbound flow speed magnitude $V$, the density $\rho$ and the dynamic viscosity $\mu$ of air. 

By combining $c_{L,\text{pre}}$ and $c_{L,\text{st}}$ with a sigmoid transition function $\sigma(\alpha, \alpha_\text{st}, M)$ \cite{beard_small_2012}, the overall three dimensional wing lift coefficient $c_{L,w}$ and drag coefficient $c_{D,w}$ across
angles of attack between -90° and 90° are determined as in (\ref{eq:aero_summed_ceofficients}). The sigmoid transition function depends on an empirically or numerically identified stall angle $\alpha_\text{st}$ and stall-transition parameter $M$ at an average sweep configuration and flight speed. These parameters are assumed constant for a specified wing profile section across flight speeds and sweep angles in our model.

\vspace{2pt}
\begin{equation}
\begin{aligned}
    c_{L,w}(\alpha, V, \Theta_w) &= (1 - \sigma)c_{L,\text{pre}} + \sigma c_{L,\text{st}} \\
    c_{D,w}(\alpha, V, \Theta_w) &= (1 - \sigma)c_{D,\text{pre}} + \sigma c_{D,\text{st}}
\label{eq:aero_summed_ceofficients}
\end{aligned}
\end{equation}
\vspace{2pt}

The center of pressure $\mathbf{r}_\text{cp}$ of a wing determines the longitudinal aerodynamic pitch moment exerted by the aerodynamic wing load. The center of pressure in the presented model lies at a chordwise location $x_\text{cp}$ from the leading edge of the wing, which ranges between the quarter- and the mid-chord position, depending on the angle of attack\cite{lindenburg_stall_2000}, as described in (\ref{eqn:wing_cp}).

\vspace{1pt}
\begin{equation}
\begin{aligned}
    & x_\text{cp} = \frac{\overline{c}(\Theta_w)}{4}(1 + \frac{2\normSI{\alpha}}{\pi}), \text{for } -\frac{\pi}{2} < \alpha < \frac{\pi}{2} \\
    & \mathbf{r}_\text{cp}(\alpha, \Theta_w) = [x_\text{cp}, 0]^\top - \mathbf{r}_\text{cg}
\label{eqn:wing_cp}
\end{aligned}
\end{equation}
\vspace{1pt}

The aircraft dynamics are additionally affected by aerodynamic damping effects typically considered for fixed-wing drones \cite{selig_real-time_2014}. These are accounted for with the quasi-steady velocity induced by the drone rotation in body frame ($\mathbf{v}_{\text{qs},b}$) at the location of the mid-chord of the wing. This quasi-steady motion $\mathbf{v_{\text{qs}}}$ is consequently combined with the freestream flow velocity to yield an inbound flow velocity, given by $\mathbf{v}_{i,b} = - (\mathbf{v}_b + \mathbf{v_{\text{qs},b}})$.

The wash induced by the propeller causes the so-called propeller slipstream, leading to an added axial flow speed $u_{s}$ inbound on portions of the wing and the tail. We apply momentum theory to obtain the  slipstream flow magnitude $u_s$ in (\ref{eqn:slipstream}). The flow velocity vector including slipstream effects in drone body-frame over the wing and tail is given as  $\mathbf{v}_{s,b} = \mathbf{v}_{i,b} - [u_{s}, 0]^\top$. This flow velocity is inbound  on a wing sub-section assumed to span the diameter of the drone propeller $2R_{p}$ and with a geometrically calculated area $S_{s}$.

\vspace{1pt}
\begin{equation}
    u_{s} = -\frac{u}{2} + \frac{1}{2} \sqrt{u^2 + \frac{2T_{m}}{\rho \pi R_{p}^2}}
    % & \mathbf{v}_{i,b} = -(\mathbf{v}_b + \mathbf{v}_{\text{qs},b}) \\
    % & \mathbf{v}_{s,b} = \mathbf{v}_{i,b} - [u_s, 0]^\top \\
\label{eqn:slipstream}
\end{equation}
\vspace{1pt}

We consequently approximate the aerodynamic load vector acting on the entire wing $\mathbf{f}_{w,b}$  by adding the aerodynamic loads acting on the wing portion within the propeller slipstream ($\mathbf{f}_{w,b}^{s}$) and the loads acting on the remaining portion ($\mathbf{f}_{w,b}^{f}$). The total longitudinal pitch moment induced by the wing ($\tau_{w,b}$) is similarly obtained by adding the individual pitch moment contributions of the two wing portions. We additionally include a static pitch moment offset term $\tau_{w,0}$ that is applicable for cambered wings and other aerodynamic pitch moments acting on the drone body. The above is formulated in (\ref{eqn:wing_body_forces}).

% \begin{equation}
% \begin{split}
%     L_{w,f} &= \tfrac{1}{2} \rho (S_{w}(\Theta_w) - S_{s}) \normDB{\mathbf{v}_{i,b}}^2 
%               c_{L,w}(\alpha_w, \normDB{\mathbf{v}_{i,b}}, \Theta_w) \\ 
%     L_{w,s} &= \tfrac{1}{2} \rho S_{s} \normDB{\mathbf{v}_{s,b}}^2 
%               c_{L,w}(\alpha_{w,s}, \normDB{\mathbf{v}_{s,b}}, \Theta_w) \\
%     D_{w,f} &= \tfrac{1}{2} \rho \big[(S_{w}(\Theta_w) - S_{s}) \normDB{\mathbf{v}_{i,b}}^2 
%               c_{D,w}(\alpha_w, \normDB{\mathbf{v}_{i,b}}, \Theta_w)\big] \\
%     D_{w,s} &= \tfrac{1}{2} \rho S_{s} \normDB{\mathbf{v}_{i,b}}^2 
%               c_{D,w}(\alpha_{w,s}, \normDB{\mathbf{v}_{s,b}}, \Theta_w) \\
%     \mathbf{f}_{w,b}^{f} &= \mathbf{R}_\alpha(\alpha_w) [D_{w}, L_{w}]^\top \\
%     \mathbf{f}_{w,b}^{s} &= \mathbf{R}_\alpha(\alpha_{w,s}) [D_{w,s}, L_{w,s}]^\top \\
%     \mathbf{f}_{w,b} &= \mathbf{f}_{w,b}^{f} + \mathbf{f}_{w,b}^{s} \\
%     \tau_{w,0} &= \tfrac{1}{2} \rho  \overline{c}(\Theta_w) c_{\tau,0}  
%               \left[(S_{w}(\Theta_w) - S_{s}) \normDB{\mathbf{v}_{i,b}}^2 
%               + S_{s} \normDB{\mathbf{v}_{s,b}}^2\right] \\
%     \tau_{w,b} &= \mathbf{r}_{\text{cp}}(\alpha_{w}, \Theta_w) \times \mathbf{f}_{w,b}^{f} 
%                + \mathbf{r}_{\text{cp}}(\alpha_{w,s}, \Theta_w) \times \mathbf{f}_{w,b}^{s} 
%                + \tau_{w,0}
% \end{split}
% \label{eqn:wing_body_forces}
% \end{equation}

\vspace{1pt}
\begin{align}
    L_{w,f} &= \frac{1}{2} \rho (S_{w}(\Theta_w) - S_{s}) \normDB{\mathbf{v}_{i,b}}^2 c_{L,w}(\alpha_w, \normDB{\mathbf{v}_{i,b}}, \Theta_w) \nonumber \\ 
    L_{w,s} &= \frac{1}{2} \rho S_{s} \normDB{\mathbf{v}_{s,b}}^2 c_{L,w}(\alpha_{w,s}, \normDB{\mathbf{v}_{s,b}}, \Theta_w) \nonumber  \\
    D_{w,f} &= \frac{1}{2} \rho [(S_{w}(\Theta_w) - S_{s}) \normDB{\mathbf{v}_{i,b}}^2 c_{D,w}(\alpha_w, \normDB{\mathbf{v}_{i,b}}, \Theta_w) \nonumber  \\
    D_{w,s} &= \frac{1}{2} \rho S_{s} \normDB{\mathbf{v}_{i,b}}^2 c_{D,w}(\alpha_{w,s}, \normDB{\mathbf{v}_{s,b}}, \Theta_w) \nonumber  \\
    \mathbf{f}_{w,b}^{f} &= \mathbf{R_{\alpha}}(\alpha_w) [D_{w}, L_{w}]^\top 
    \label{eqn:wing_body_forces}\\
    \mathbf{f}_{w,b}^{s} &= \mathbf{R_{\alpha}}(\alpha_{w,s}) [D_{w,s}, L_{w,s}]^\top \nonumber  \\
    \mathbf{f}_{w,b} &= \mathbf{f}_{w,b}^{f} + \mathbf{f}_{w,b}^{s} \nonumber  \\
    \tau_{w,0} & = \frac{1}{2} \rho  \overline{c}(\Theta_w) c_{\tau,0}  \left[(S_{w}(\Theta_w) - S_{s}) \normDB{\mathbf{v}_{i,b}}^2 + S_{s} \normDB{\mathbf{v}_{s,b}}^2\right]  \nonumber  \\
    \tau_{w,b} &= \mathbf{r}_{\text{cp}}(\alpha_{w}, \Theta_w) \times \mathbf{f}_{w,b}^{f} + \mathbf{r}_{\text{cp}}(\alpha_{w,s}, \Theta_w) \times \mathbf{f}_{w,b}^{s} + \tau_{w,0} \nonumber
\nonumber
\end{align}
\vspace{1pt}

where, assuming a symmetric wing and no other added aerodynamic static pitch moments for the presented prototype, $c_{\tau,0} = 0 $. The angles of attack in the slipstream region and outside of the slipstream region are denoted as $\alpha_{w,s}$ and $\alpha_w$ respectively. The rotation matrix $\mathbf{R}_\alpha \in \mathbb{R}^{2\times 2}$ transforms the aerodynamic loads from the wind frame to the drone body frame. 

Similar considerations are made to model the aerodynamic loads acting on the drone tail. The slipstream velocity term defined in (\ref{eqn:wing_body_forces}) is applied to model the tail lift and drag forces. Flat-plate aerodynamic coefficients \cite{cory_experiments_2008} evaluated at an effective tail angle of attack $\alpha_{t,\text{eff}} = k_\text{ele}\alpha_{t}$ are used to derive the tail force vector, where $\alpha_{t}$ is the angle of attack of the flow inbound on the tail including the propeller slipstream. The elevator effectiveness factor $k_\text{ele}$ is derived from the surface area ratio of the elevator to the entire horizontal tail  \cite{nelson_flight_1998}. The tail pitch moment $\tau_{t,b}$ is obtained considering a tail-force vector acting at the horizontal tail mid-chord location.

\begin{figure*}
    \centering
    \includegraphics[width=\linewidth]{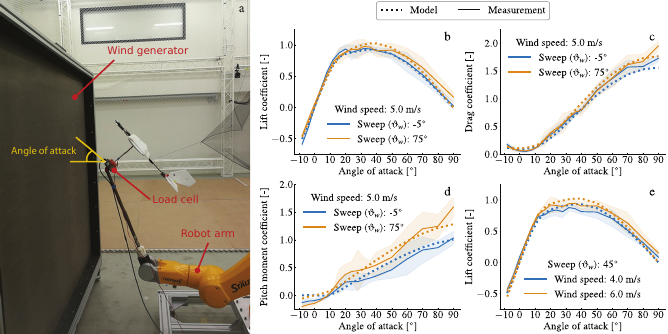}
    \caption{\textbf{(a)} Aerodynamic characterization setup using ATI Gamma load cell, a wind generator and a robot arm positioning the drone at angles of attack between -8° and 90°. We compare the derived aerodynamic model to measurements of \textbf{(b)} lift coefficients, \textbf{(c)} drag coefficients, and \textbf{(d)} pitch moment coefficients  at a sweep angle of -5° (extended wing configuration) and 75° (fully swept wing configuration) at an inbound wind speed of 5 m/s. \textbf{(e)} Lift coefficient measurements and aerodynamic model predictions at wind speeds of 4 and 6 m/s at a wing sweep angle of 45° are additionally shown.}
    \label{fig:characterization}
\end{figure*}

The static aerodynamic force and torque model predictions were experimentally validated on the drone prototype in a wind-tunnel setup. The drone was placed on a robotic arm (\textit{Stäubli TX-90}) that could vary the angle of attack between 0-90°, in steps of 4° between 0° and 40° and in steps of 10° between 50° and 90°. An open wind tunnel (\textit{WindShape}) generated air speeds of 4 - 6 m/s.  The lift and drag forces were obtained using a 6-axis force/torque sensor (\textit{ATI Gamma SI-130-10}) attached between the drone and the robotic arm  [Fig. \ref{fig:characterization}(a)]. We compare the performance of the aerodynamic model with the measured lift, drag and pitch moment coefficients at a wind speed of 5 m/s [Fig. \ref{fig:characterization}(b)-(d)]. The model captures the post-stall regime, which occurs above an angle of attack of $\sim$ 14°, within one standard deviation of the measurements. A lower lift-curve slope in fully-swept (75°) configuration compared to the extended (-5°) configuration is furthermore evidenced both in the model and from measurement [Fig. \ref{fig:characterization}(a)].  It is seen that the pitch moment at pre-stall negative angles of attack is narrowly underestimated compared to the measurement. Origins may be differences in unmodeled wing-induced downwash experienced by the tail at negative angles of attack [Fig. \ref{fig:characterization}(d)]. The measured and modeled lift-coefficient at a mid-sweep morphing configuration and wind speeds of 4 ($\text{Re} = 45000$) and 6 m/s ($\text{Re} = 68000$) shows the aforementioned low-Reynolds lift degradation [Fig. \ref{fig:characterization}(e)]. 

\section{Optimal gap passage trajectories} %(2-3 pages)
\label{sec:traj_opt_formulation}

\begin{figure*}
    \centering
    \includegraphics[width=\linewidth]{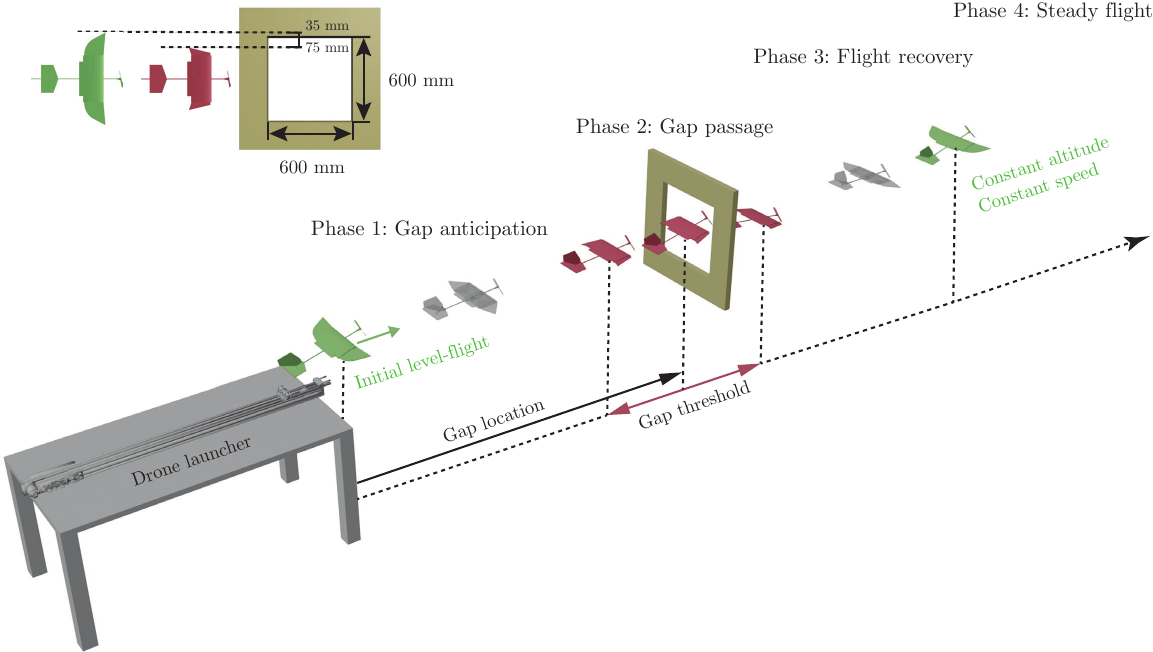}
    \caption{Description of the morphing-winged drone narrow gap passage optimization problem: The drone approaches from level flight at a specified initial flight speed. During a chosen gap anticipation distance, the drone can vary its symmetric wing sweep, thrust and elevator inputs freely to preempt the narrow-gap passage (Phase 1). During a specified gap threshold distance, the symmetric wing sweep must remain in a fully swept actuator state and the drone must maintain an altitude corresponding to the center of the narrow gap (Phase 2). After gap passage, the drone is given a free distance to recover to a constant flight condition at the altitude corresponding to the gap center, with free choice of symmetric-wing sweep, elevator and thrust (Phase 3). In the final phase (Phase 4), the drone resumes steady flight at the initial launch altitude with a freely chosen constant velocity, attitude and wing sweep configuration.}
    \label{fig:trajectory_overview}
\end{figure*}

\begin{figure*}
    \centering
    \includegraphics[width=\linewidth]{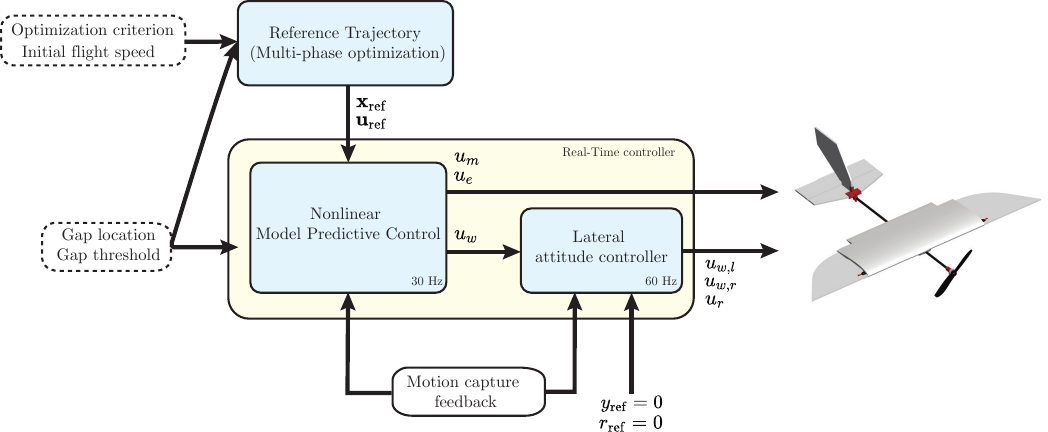}
    \caption{Optimization framework and control architecture, showing the generation of the reference trajectory using multi-phase optimization, the Nonlinear Model Predictive controller to generate symmetric sweep, motor thrust and elevator servo commands, and the low-level lateral attitude controller computing asymmetric wing sweep and rudder corrections to maintain lateral alignment to the gap.}
    \label{fig:architecture}
\end{figure*}

Observations of budgerigars flying through a narrow gap suggest that birds anticipate an upcoming gap that is narrower than their extended wingspan by adjusting their flight altitude\cite{vo_anticipatory_2016}. The birds fully tuck their wings when flying through a gap of sub-wingspan width to prevent a collision, regardless of the gap width\cite{schiffner_minding_2014}. 

Similarly, a drone should automatically time wing sweeping to prevent collision of the wings with the gap and adjust its posture ahead of the gap to preempt the loss of lift while the wings are fully swept. Here, we use multi-phase trajectory optimization methods to generate a reference trajectory considering  the aerodynamic model described above and the constraints imposed by the narrow gap. We then utilize Nonlinear MPC to track these trajectories and investigate gap passage accuracy on the real-drone.

We structure the gap passage problem and experimental setup as shown in Fig. \ref{fig:trajectory_overview}. We assume an initial velocity corresponding to level flight with fully extended wings, a specified gap location and gap threshold. The gap threshold defines the distance during which the wings remain fully swept in order to provide tolerance for control errors and avoid wingtip collision with the lateral edges of the gap  [see Fig. \ref{fig:trajectory_overview}]. The entire maneuver consists of four phases: \textbf{(1)} gap anticipation, \textbf{(2)} gap passage, \textbf{(3)} recovery to steady flight, and \textbf{(4)} steady flight. To generate optimal reference trajectories for this maneuver with maximum flexibility for phase-wise variable constraints and phase-dependent cost functions, we use multi-phase optimization based on pseudo-spectral collocation methods. We generate trajectories under combinations of three different optimization criteria, three initial flight speeds, three gap locations, and three gap thresholds.

The multi-phase optimization problem is formulated in the form of (\ref{eqn:mpopt}), considering the states, inputs and dynamics $\mathbf{f}(\mathbf{x}, \mathbf{u})$ described in (\ref{eqn:dynamics_eom}). Inequality constraints $g(\mathbf{x}, \mathbf{u}, t, p)$, and equality constraints $h(\mathbf{x}_0, t_0, \mathbf{x}_f, t_f, p)$ are defined individually for each of the four phases $p \in 	\{0, 1, 2, 3 \}$. The implementation is made with the publicly available package MPOPT. \cite{thammisetty_development_2020}.

\vspace{1pt}
\begin{align}
\label{eqn:mpopt}
    &\min_{\mathbf{x}, \mathbf{u}, t_0, t_f, p} & \hspace{2pt} & J = L_0 (\mathbf{x}_0, t_0,\mathbf{x}_f, t_f, p) + \int_{0}^{t_f}L(\mathbf{x}, \mathbf{u}, t, p)dt \nonumber\\
    &\text{subject to} &      & \dot{\mathbf{x}} = \mathbf{f}(\mathbf{x}, \mathbf{u}) \nonumber \\
    &                  &      & g(\mathbf{x}, \mathbf{u}, t, p) \leq 0   \\
    &                  &      & h(\mathbf{x}_0, t_0, \mathbf{x}_f, t_f, p) = 0 \nonumber
\end{align}
\vspace{1pt}

We specify three different optimization criteria by varying the phase-dependent elements $L_0 (\mathbf{x}_0, t_0, \mathbf{x}_f, t_f, p)$ and $L(\mathbf{x}, \mathbf{u}, t, p)$ of the cost function in order to assess their impact on the resulting reference trajectories and the drone maneuvers on hardware. The first criterion (Case 1) consists of minimizing altitude variation throughout the entire maneuver. The second criterion (Case 2) consists of minimizing the variation in forward flight speed throughout the entire maneuver. This criterion is selected to assess whether constrained speed variability still permits successful gap traversal by the morphing-winged drone. The third criterion (Case 3) consists of minimizing the duration for which the wings remain fully swept during gap passage. This strategy is inspired by the hypothesis that birds fold their wings only when necessary to minimize altitude-loss during gap passage, based on observations of budgerigars \cite{schiffner_minding_2014}. We  study whether this phenomenon translates to the morphing-drone by leading to a reduction in altitude error at the gap location and during the flight recovery phase after gap-passage.

\section{Flight control architecture} %(1 page)

A control architecture is designed to track the optimal longitudinal trajectories, maintain lateral alignment to the gap, and enforce collision safety by fully sweeping the wings across the gap threshold distance on hardware. Previously, cascaded Proportional-Integral-Derivative (PID) controllers have been used to maintain steady banking turns with coordinated ailerons and sweeping wings \cite{liu_employing_2023}. A PID control strategy for outdoor loitering flight with morphing wings and tail was employed to control the feathered PigeonBot II \cite{chang_bird-inspired_2024}. Similarly, a cascaded PID controller was used to steadily hover against airflow and reject gust disturbances \cite{jeger_adaptive_2024}. 

However, PID control lacks predictive capabilities and can therefore not preemptively adjust the drone control inputs in the presence of wing sweeping constraints imposed by the narrow gap. To overcome this limitation, we propose a Nonlinear Model Predictive Control (MPC) framework for accurate reference trajectory tracking in flight through narrow gaps. By incorporating the full nonlinear drone dynamics model including actuator dynamics, our controller inherently predicts when wing sweeping must be initiated at runtime, ensuring that the fully swept-wing constraints during the gap passage phase are satisfied and collisions are avoided. Prior work applied Nonlinear MPC to morphing-wing perching maneuvers \cite{wuest_agile_2024} at a fixed initial speed, but did not use thrust actuation or accounted for time-varying costs and actuator limits.

In our control architecture [Fig. \ref{fig:architecture}], MPC solutions are computed at an update rate of 30 Hz to regulate elevator, throttle and symmetric wing sweep commands. In parallel, a low-level, cascaded PID controller with adaptive actuator allocation regulates asymmetric wing sweep and rudder at 60 Hz to compensate for lateral deviations from center of the gap. The Nonlinear MPC problem is implemented offboard using multiple shooting and solved with the High Performance Interior Point Method (HPIPM) solvers in the ACADOS Toolkit \cite{verschueren_acados_2020}. The nonlinear dynamics (\ref{eqn:dynamics_eom}) are integrated using Runge-Kutta (RK4) to describe the dynamics constraints in the form of a set of functions $\mathbf{f}_{rk4}$ that depend on the state $\mathbf{x}_k$ and actuation input $\mathbf{u}_k$ at every shooting stage $k$. The optimization problem is initialized with a state vector $\hat{\mathbf{x}}_0$, which concatenates the observed longitudinal state and estimated actuator state derived from the actuator dynamics models (\ref{eqn:dynamics_eom}). We select the state and actuator input references $\mathbf{x}_{\text{ref},k}$ and $\mathbf{u}_{\text{ref},k}$ from one of the reference trajectories (Section \ref{sec:traj_opt_formulation}), identifying the first reference point as that closest to the presently measured position. At every step of the MPC control loop, we set the prediction horizon $k_f$ to the remaining number of reference trajectory points under the condition that $k_f \leq N$, where $N$ is a pre-determined maximum prediction horizon. Consequently, the number of shooting stages is given as $k \in [0, k_f]$. Furthermore, based on the gap location $x_\text{gap}$ and the gap threshold distance $x_\text{thr}$, within every control execution step, we identify the set of shooting stages $k_{\text{gap}} \in [k_{\text{gap},i}, k_{\text{gap},f}]$ at which the reference drone position, denoted by $x_\text{ref}$, lies within the bounds of the gap passage threshold, $x_{\text{ref},k_{\text{gap}}} \in [x_\text{gap} - x_\text{thr}/2, x_\text{gap} + x_\text{thr}/2]$. To enforce a fully swept wing actuator input ($u_\text{w}$) and corresponding actuator state ($x_\text{w}$) for the set of shooting stages $k_{\text{gap}}$ within the gap threshold distance, we formulate linear state and input constraints, $\mathbf{x} \in [\mathbf{x}_{\text{min},g}, \mathbf{x}_{\text{max},g}]$ and $\mathbf{u} \in [\mathbf{u}_{\text{min},g}, \mathbf{u}_{\text{max},g}]$ respectively. The nominal actuator and state limits outside of this region are $\mathbf{x} \in [\mathbf{x}_{\text{min},0}, \mathbf{x}_{\text{max},0}]$ and $\mathbf{u} \in [\mathbf{u}_{\text{min},0}, \mathbf{u}_{\text{max},0}]$. We further define a shooting-stage dependent quadratic cost to prioritize altitude accuracy and guide wing sweeping at the shooting stage within the gap threshold. The optimal control problem is formally defined in (\ref{eqn:mpc_base}) for shooting stages during the gap passage phase and outside of the gap passage phase.

% ($\mathbf{q}_g$ and $\mathbf{r}_g$)
% ($\mathbf{q}_0$ and $\mathbf{r}_0$).

\begin{equation}
\begin{aligned}
&\min_{\mathbf{x}_{[1:k_{f}]}, \mathbf{u}_{[0:k_f-1]}} & & \sum_{k=1}^{k_f} (\mathbf{x}_{\text{ref},k} - \mathbf{x}_k)^\top \mathbf{Q} (\mathbf{x}_{\text{ref},k} - \mathbf{x}_k) + \ldots \\
&& & \ldots \hspace{2pt} + \sum_{k=0}^{k_f-1} (\mathbf{u}_\text{ref,k} - \mathbf{u}_k)^\top \mathbf{R}  (\mathbf{u}_{\text{ref},k} - \mathbf{u}_k) \\
&\text{subject to:}&      & {\mathbf{x}}_{k+1} = \mathbf{f}_\text{rk4}(\mathbf{x}_k, \mathbf{u}_k) \\
&                  &      & \mathbf{x}_0 = \hat{\mathbf{x}}_0 \\
&                  &      & \mathbf{x}_{\text{min},0} \leq \mathbf{x}_k \leq \mathbf{x}_{\text{max},0} \hspace{3pt} ,\text{if } k \notin [k_{\text{gap},i}, k_{\text{gap},f}] \\
&                  &      & \mathbf{x}_{\text{min},g} \leq \mathbf{x}_k \leq \mathbf{x}_{\text{max},g} \hspace{3pt} ,\text{if } k \in [k_{\text{gap},i}, k_{\text{gap},f}] \\
&                  &      & \mathbf{u}_{\text{min},0} \leq \mathbf{u}_k \leq \mathbf{u}_{\text{max},0} \hspace{3pt} ,\text{if } k \notin [k_{\text{gap},i}, k_{\text{gap},f}] \\
&                  &      & \mathbf{u}_{\text{min},g} \leq \mathbf{u}_k \leq \mathbf{u}_{\text{max},g} \hspace{3pt} ,\text{if } k \in [k_{\text{gap},i}, k_{\text{gap},f}] \\
&                  &      & \mathbf{Q} = \text{diag}(\mathbf{q}) \\
&                  &      & \mathbf{R} = \text{diag}(\mathbf{r}) \\
\end{aligned}
\label{eqn:mpc_base}
\end{equation}

\text{where:}

\begin{align}
& \mathbf{q} = \begin{cases}
 [1, 100, 1, 1, 0, 0, 0, 2, 0, 0]^\top & \text{if } k \notin [k_{\text{gap},i}, k_{\text{gap},f}], \nonumber \\
 [1, 1000, 1, 1, 0, 0, 0, 100, 0, 0]^\top  & \text{if } k \in [k_{\text{gap},i}, k_{\text{gap},f}]. \nonumber \\
\end{cases} \nonumber \\
& \mathbf{r} = \begin{cases}
 [0, 0, 2]^\top & \text{if } k \notin [k_{\text{gap},i}, k_{\text{gap},f}], \nonumber \\
 [0, 0, 100]^\top  & \text{if } k \in [k_{\text{gap},i}, k_{\text{gap},f}]. \nonumber\\
\end{cases}
\nonumber
\end{align}

In closed-loop operation, position and orientation feedback is provided by a motion capture system, and translational and angular velocity states are derived using a first-order Butterworth filter. We account for delays introduced by communication latency between the offboard computer and the onboard microcontroller, computation time, and actuator lag. We do so by propagating the system dynamics forward in time by a constant delay using prior input history and the instantaneous measured state. This delay-compensated state prediction is provided to the Nonlinear MPC solver. The combined delays amounted to 65 ms when measured with offboard control on our prototype. The Nonlinear MPC and lateral PID are executed on a laptop with an Intel i9-12900H CPU. We set the maximum prediction horizon to $N = 30$ (1.0 s), leading to a $\sim$ 6 ms execution time per MPC control timestep. 

The low-level lateral PID controller is commanded to maintain a fixed lateral (y-axis) position aligned with the gap center ($y_\text{ref}$). In a cascaded structure, a reference roll angle, body frame roll rate and roll acceleration are determined. A zero yaw rate ($r_\text{ref}$) is commanded, and a PID determines the reference yaw acceleration  [Fig. \ref{fig:architecture}]. The roll and yaw acceleration commands are consequently converted to action commands for the rudder ($u_r$) and asymmetric sweep offsets ($\Delta u_\text{w}$) using a state-dependent actuator allocation matrix. Left and right wing sweep inputs are  computed as $u_{w,l} = u_{w} + \Delta u_{w}$ and $u_{w,r} = u_{w} - \Delta u_{w}$ respectively. When the wings are symmetrically swept in by the higher level MPC during the gap passage phase, no asymmetric sweep corrections are performed.  A simplified flat-plate aerodynamics model is used to determine analytical expressions for the effectiveness of each control surface in yaw and roll, thereby accounting for cross-coupling effects between the actuators that vary with drone velocity, angle of attack, and angle of sideslip.  These expressions are computed in real-time from the measured drone velocity vector and define the entries for the actuator allocation matrix. 

\section{Results}

\subsection{Trajectory optimization objectives affect flight maneuvers}

\begin{figure}[h]
    \includegraphics[width=\columnwidth]{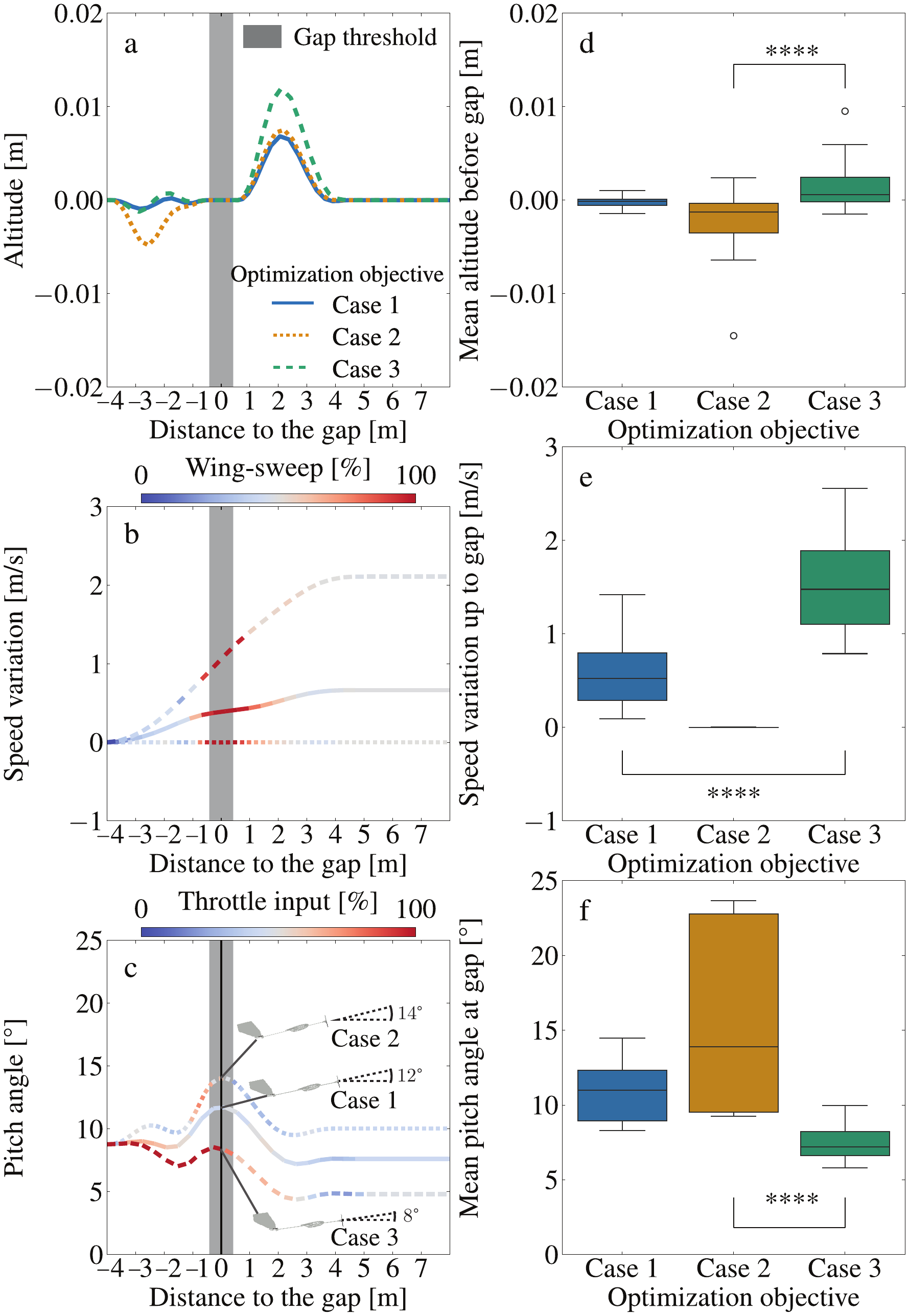}
    \caption{Comparison of optimal reference trajectories among three flight optimization objectives: minimum variation of altitude across all flight phases (Case 1), minimum variation in forward flight speed (Case 2), and minimum time during the gap passage phase (Case 3). For the exemplary case of an initial level flight speed of 6 m/s, a gap location at 4 m, and a gap threshold of 0.8 m, we show: \textbf{(a)} Altitude relative to the launch altitude, \textbf{(b)} flight speed relative to approach speed with wing sweep input overlay (100 \% represents fully swept wings), and \textbf{(c)} drone pitch angle with throttle input overlay. Boxplots display the the following for every flight optimization objective summarized across three initial level flight speeds (5,6 and 7 m/s), gap thresholds (0.4, 0.8 and 1.2 m) and gap locations (4, 5 and 6 m): \textbf{(d)} Time-averaged altitude during the gap anticipation phase, \textbf{(e)} Gap passage flight speed relative to approach speed, \textbf{(f)} Time-averaged pitch angle during the gap passage phase. A Mann-Whitney U-test was performed between datasets of interest, yielding P $<$ 0.0001 (****) between all compared selections.}
    \label{fig:traj_case_comparison}
\end{figure}

We generate trajectories for all combinations across three different gap locations (4, 5 and 6 m from the launch point), three different gap thresholds (0.4, 0.8 and 1.2 m), three different initial speeds (5, 6 and 7 m/s) and three different trajectory optimization objectives (Cases 1-3). Across all flight conditions, we specifically study how the optimal drone behavior changes ahead of the narrow gap depending the choice of optimization objective.

For all objectives and all tested flight conditions, the average altitude variation ahead of the gap is less than $\sim$ 1.5 cm [Fig. \ref{fig:traj_case_comparison}(a) and (d)]. The altitude tends to decrease when the forward velocity remains constant (Case 2) and increases when the time of fully swept wings is minimized (Case 3) [see Fig. \ref{fig:traj_case_comparison} (d)] (Mann-Whitney U-Test, n=27, p $<$ 0.0001). However, in relation to the several meters of flight distance covered ahead of the narrow gap passage phase, these differences are negligible. The extent to which flight speed is gained  ahead of the narrow gap varies between the three optimization objectives [Fig. \ref{fig:traj_case_comparison} (b)]. The median flight speed in Case 3 (minimum time spent in the gap passage) is $\sim$ 1 m/s higher than that in Case 1 (minimal variation in altitude) [see Fig. \ref{fig:traj_case_comparison}(e)] (Mann-Whitney U-Test, n=27, p $<$ 0.0001). The wing sweep percentages show progressive sweep-in ahead of the gap and recovery to a steady value during the recovery phase after gap-passage, with consistent behavior across the three trajectory optimization objectives [Fig. \ref{fig:traj_case_comparison} (b)]. 
We furthermore observe a pitch-up maneuver ahead of the gap with an inflection point at the gap location, followed by a pitch-down after gap passage to recover to level flight [Fig. \ref{fig:traj_case_comparison}(c)]. The pitch angle at the gap location and the applied throttle ahead of the gap vary between the three studied trajectory optimization objectives [Fig. \ref{fig:traj_case_comparison}(c)]. Across all tested flight conditions, the trajectories reveal strong evidence that the pitch-angle attained at gap passage depends on the flight objective. The pitch-angle at the gap location is highest for Case 2 (minimal variation in forward speed) and lowest for Case 3 (minimal time with fully swept wings) [Fig. \ref{fig:traj_case_comparison}(f)] (Mann-Whitney U-Test, n=27, p $<$ 0.0001).

In summary, our analysis of the generated trajectories shows that the choice of trajectory optimization objective influences the mechanisms the drone employs to anticipate narrow-gap passage. Optimizing for a constant forward flight speed ahead, during and after gap passage (Case 2) leads to a larger increase in pitch angle ahead of the gap. On the contrary, minimizing the time during which the wings are fully swept (Case 3) leads to a larger use of thrust to gain speed up to the gap and a lower pitch angle at the gap location. However, despite the lowest average change in altitude ahead of the gap being observed when the altitude variation across the trajectory is to be minimized (Case 1), the altitude remains similarly constant - varying overall by less than 1.5 cm prior to the gap - when the other optimization objectives are employed. 

\begin{figure*}[ht]
    \centering
    \includegraphics[width=\linewidth]{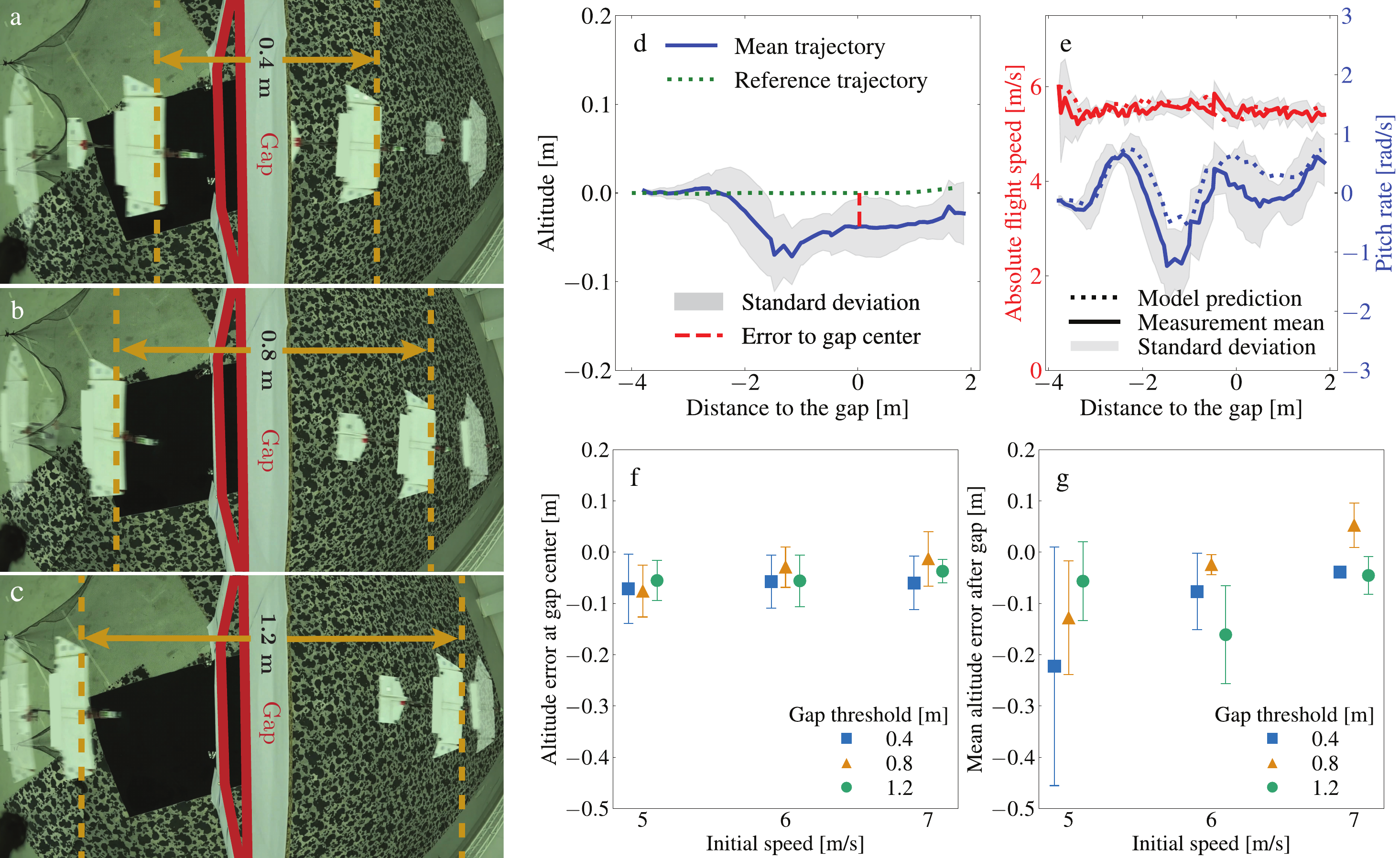}
    \caption{Narrow-gap passage flight experiments for closed-loop control with Nonlinear MPC and reference trajectories for a flight optimization objective of least altitude variation (Case 1). We visually show the gap passage phase with fully swept wings and the inward sweeping and extending process for an initial speed of 6 m/s and a gap threshold of \textbf{(a)} 0.4 m, \textbf{(b)} 0.8 m and \textbf{(c)} 1.2 m. Exemplified considering an experiment with an initial speed of 6 m/s and a gap threshold of 0.8 m, we show \textbf{(d)} the barycentric mean trajectory across three flights compared to the computed reference trajectory with the computed vertical error to the gap center and \textbf{(e)} the predicted mean absolute flight speed and mean pitch rate by the dynamic model over one MPC prediction time-step compared to values measured across three flight experiments. For flight experiments between initial speeds of 5, 6 and 7 m/s and gap thresholds varying between 0.4, 0.8 and 1.2 m, we compare the \textbf{(f)} Altitude error measured at the gap location, and the \textbf{(g)} Mean altitude tracking error during the recovery phase.}
    \label{fig:flight_experiment_fig_1}
\end{figure*}

\subsection{Flight experiments show collision-free gap passage}

We conducted closed-loop control flight experiments in a 10 x 10 x 8 m flight arena to determine the altitude error at and after gap passage across variable initial speeds and gap thresholds. Additionally, we studied how these errors are affected by utilizing reference trajectories under the different optimization objectives described in the previous section. We launched the drone at desired speeds of 5, 6 and 7 m/s by means of an automated linear launcher [Fig. 4]. We positioned the gap at a distance of 4 m from the launch point and collected measurement data for a total distance of 6 m from the launch point. We conducted three flight experiments for each combination of gap threshold, initial flight speed, and optimization objective. For each set of three flights, we computed the mean trajectory and the corresponding standard deviations for position, pitch angle, pitch rate, and velocity. For all conditions tested, we also determined the Root-Mean-Square-Error (RMSE) between the full reference and measured flight trajectories [Table \ref{tab:rmse_trajectory_table}].

\subsubsection{Varying initial speed and gap threshold}

For hardware deployment with closed-loop MPC, the initial approach speed to the gap and the gap threshold are important factors determining the accuracy with which the controller can achieve successful gap passage. In particular, at a lower speed and at higher gap thresholds, the controllers ability to compensate lift-loss during gap passage is tested, whilst at higher speeds and at lower gap thresholds, the demands for timely initiation of inward wing sweeping to avoid collision pose significant control challenges. We therefore evaluate the gap passage closed-loop control accuracy in altitude for combinations of gap thresholds of 0.4, 0.8 and 1.2 m and initial flight speeds of 5, 6 and 7 m/s, assuming a constant trajectory optimization objective minimizing the variation in altitude (case 1). 

At all three flight speeds, the experiments showed that the drone successfully initiated wing sweeping before gap passage and extended its wings afterward at different locations specified by the gap threshold [Fig. \ref{fig:flight_experiment_fig_1}(a) to (c), see supplementary video]. Tested flight speed averaged gap passage phase times with fully swept wings ranged between 55 ms and 262 ms. Ahead of the gap, the wings remained on average $\sim$ 45 $\%$ (SD = 4.4 $\%$) swept. The absolute lateral error to the center of the gap was evaluated as 2.8 cm (SD = 2.0 cm). 

We show the measured altitude trajectory compared to the reference trajectory for an example case [Fig. \ref{fig:flight_experiment_fig_1} (d)]. For all experiment cases, we extract the altitude error to the gap center and the altitude tracking error after gap-passage for control performance analysis. Considering the example case, we check the predictive capabilities of the dynamic model used within the MPC controller in-flight. To do so, from the barycentric mean measurements, the absolute flight speed and pitch rate were measured and compared to the instantaneously predicted values obtained by model propagation over one MPC control time-step. It is seen that the velocity predictions followed the trends of the measured states, yet a positive offset remains in the prediction of the pitch rate compared to the measurements [see example in Fig. \ref{fig:flight_experiment_fig_1} (e)]. 
Across the three tested initial speeds and gap thresholds, the drone achieved a mean altitude gap passage error of -5.1 cm (SD = 4.4 cm) in closed-loop MPC operation [Fig. \ref{fig:flight_experiment_fig_1} (f)]. Averaged across values for all gap thresholds at one speed, at initial speeds of 5 m/s, the mean altitude gap passage error is -6.7 cm (SD = 4.4 cm), increasing to -4.8 cm (SD = 4.0 cm) at 6 m/s and to -3.6 cm (SD = 4.1 cm) at 7 m/s [see Fig. \ref{fig:flight_experiment_fig_1} (f)]. Furthermore, altitude recovery after gap passage does not show a clear correlation to the choice of gap threshold distance valid across all flight speeds. However, with increasing speed, averaged across all gap threshold values, the altitude after gap passage also increases from a mean of -13.6 cm (SD = 14.4 cm) at an initial speed of 5 m/s, to -8.7 cm (SD = 8.1 cm) at 6 m/s and -1.1 cm (SD = 5.2 cm) at an initial speed of 7 m/s [Fig. \ref{fig:flight_experiment_fig_1} (g)].

\subsubsection{Varying the optimization objective}

\begin{figure}
    \includegraphics[width=\columnwidth]{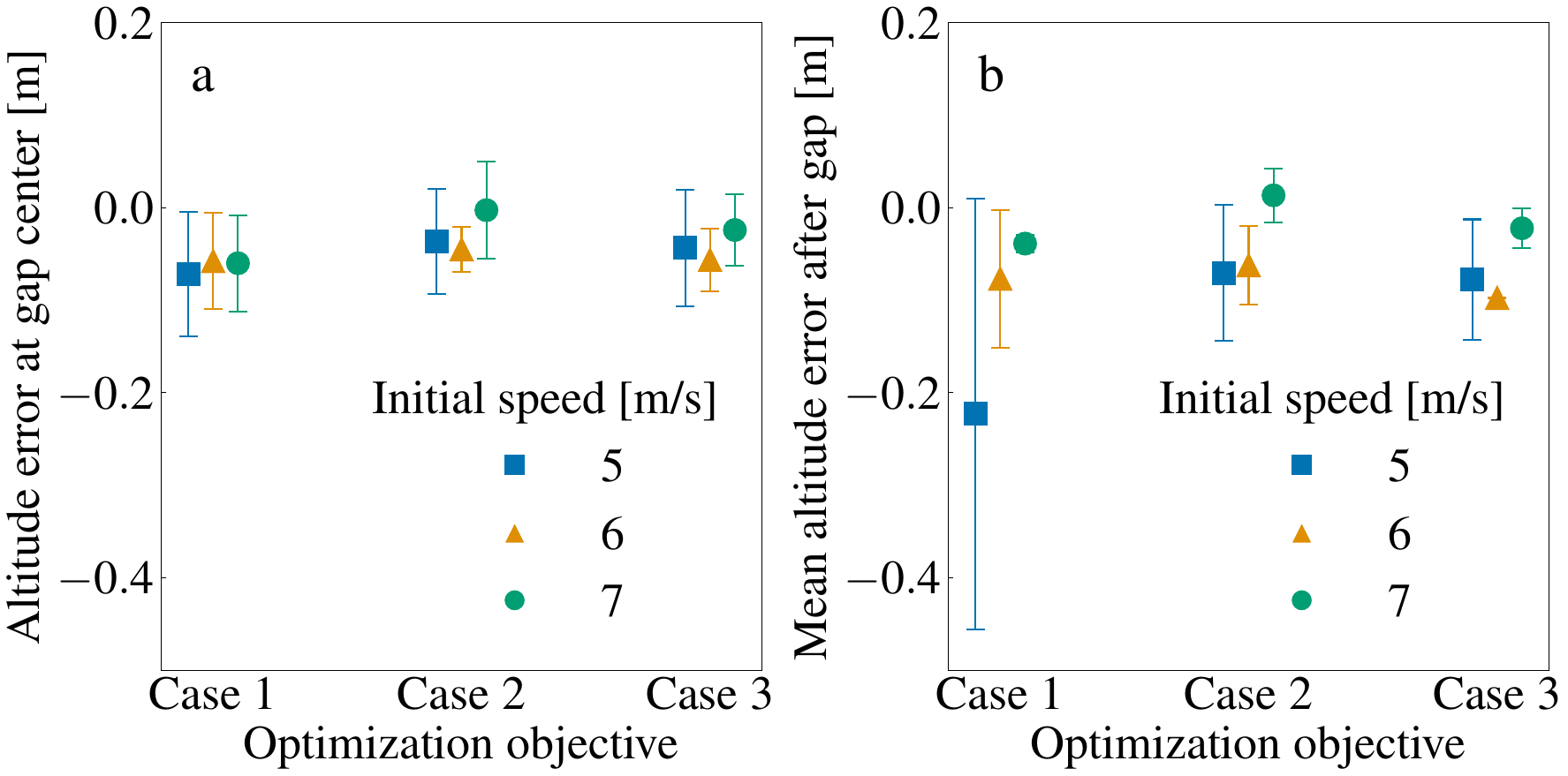}
    \caption{ Closed-loop control flight experiment data obtained for reference trajectory optimization objectives of minimum variation of altitude across all flight phases (Case 1), minimum variation in forward flight speed (Case 2), and minimum time during the gap passage phase (Case 3). We show the \textbf{(a)} altitude error measured at the gap location and \textbf{(b)} mean altitude tracking error during the recovery phase. A gap threshold of 0.4 m and initial speeds of 5, 6 and 7 m/s are set.}
    % \textbf{(g)} Snapshots of the drone at the instant of gap passage at an initial speed of 5 m/s and a gap threshold of 0.4 m highlight the variation in pitch-angle of the \textit{LISparrow} between different optimization objectives when tracked by the MPC controller.}
    \label{fig:traj_case_comparison}
\end{figure}

\begin{table}
\centering
\caption{Altitude tracking Root-mean square error (RMSE)}
% Please add the following required packages to your document preamble:
% \usepackage{multirow}
\begin{tabular}{c|ccccc|l}
\cline{2-6}
\multicolumn{1}{l|}{}                                                                                               &
\multicolumn{5}{c|}{\cellcolor[HTML]{EFEFEF}\textbf{Mean trajectory tracking altitude error {[}m{]}}}                                             &  \\ \cline{1-6}
\multicolumn{1}{|c|}{}                                                                                              & \multicolumn{5}{c|}{\textbf{Gap threshold}}                                                                                       &  \\ \cline{2-6}
\multicolumn{1}{|c|}{}                                                                                              & \multicolumn{3}{c|}{\textbf{0.4 m}}                                      & \multicolumn{1}{c|}{\textbf{0.8 m}}  & \textbf{1.2 m}  &  \\
\multicolumn{1}{|c|}{\multirow{-3}{*}{\textbf{\begin{tabular}[c]{@{}c@{}}Initial speed \\ {[}m/s{]}\end{tabular}}}} & \textit{Case 1} & \textit{Case 2} & \multicolumn{1}{c|}{\textit{Case 3}} & \multicolumn{1}{c|}{\textit{Case 1}} & \textit{Case 1} &  \\ \cline{1-6}
\multicolumn{1}{|c|}{5}                                                                                             & 0.145           & 0.047           & \multicolumn{1}{c|}{0.055}           & \multicolumn{1}{c|}{0.088}           & 0.055           &  \\
\multicolumn{1}{|c|}{6}                                                                                             & 0.050           & 0.040            & \multicolumn{1}{c|}{0.047}           & \multicolumn{1}{c|}{0.033}           & 0.068           &  \\
\multicolumn{1}{|c|}{7}                                                                                             & 0.047              & 0.039              & \multicolumn{1}{c|}{0.039}     & \multicolumn{1}{c|}{0.033}              & 0.033        &  \\ \cline{1-6}
\end{tabular}
\label{tab:rmse_trajectory_table}
\end{table}

When tracking reference trajectories generated for different optimization objectives, real flight experiments with Model Predictive Control reveal marginal differences in altitude tracking performance between the three optimization objectives. We analyzed experiments for reference trajectories generated for a constant gap-threshold of 0.4 m, at three different initial approach speeds (5-7 m/s) under three optimization objectives (Cases 1-3). 

For trajectories optimized for minimum altitude variation (Case 1), the vehicle passes the gap with a mean altitude error of -6.3 cm (SD = 4.7 cm). Under the minimum speed variation objective (Case 2), the deviation is reduced to -2.8 cm (SD = 4.2 cm), while the minimum-time objective with fully swept wings (Case 3) yields an intermediate deviation of -4.1 cm (SD = 4.0 cm) [Fig. \ref{fig:traj_case_comparison}(a)]. The same ranking is observed after gap passage: Case 1 shows the largest altitude error compared to the reference altitude (-11.3 cm, SD = 14.0 cm), Case 2 the smallest (-4.0 cm, SD = 5.7 cm) loss, and Case 3 an intermediate loss (-6.6 cm, SD = 4.6 cm), and [Fig. \ref{fig:traj_case_comparison}(b)].

\section{Conclusion}

The results described above show that the optimal reference trajectories display a variation in altitude ahead of the gap of $<$ 2 cm, independently of the trajectory optimization criteria. Instead, budgerigars traversing narrow gaps at similar approach speed ($\sim$ 4-5 m/s) display a substantially larger altitude variation  \cite{altshuler_comparison_2018}. This discrepancy can be explained by the fact that the propeller-driven morphing drone can leverage upwards thrust-vectoring at the gap location to compensate for loss of lift when wings are swept in, whilst flapping-wing birds cannot do that and must gain greater height ahead of the gap in order to compensate for inevitable lift and altitude loss caused by wing sweeping.

In flight experiments, the drone achieved collision-free flights across the gap from initial speeds as low as 5 m/s, initiating wing-sweeping 0.2 meters ahead of the gap, with an average error in altitude to the gap center of $\sim$5~cm. Experiments show that the choice of optimization objective for the reference trajectory marginally impacts the altitude tracking performance - indicating that neither minimizing changes in altitude nor minimizing the time of fully sweeping in the wings provide a significant advantage. However, optimizing for a constant forward flight speed, allows for marginally better altitude tracking performance ($\sim$ 1.4 cm) than when minimizing the time during which the wings are fully-swept. A likely explanation is that a steady speed produces a steady airflow that better matches the conditions in which the aerodynamic model was validated; this would make the model uncertainty lower and lead to better MPC trajectory tracking performance.

In these experiments with the real drone, we observed spurious variations in the roll angle between individual experiments that result in a smaller lift force and thus an altitude loss that is larger than under the level-wing assumption considered by MPC. This leads to the higher variability in trajectory tracking performance observed across repeated flight experiments. A root cause of these roll angle variations is the relatively lower effectiveness of asymmetric wing sweep at low angles of attack. A solution to better stabilize lateral attitude of the drone at low angles of attack consists in using an asymmetric wing-twist mechanism  \cite{ajanic_sharp_2022}, which the drone used here does not have.
 
Model prediction analysis showed an underestimation of pitching dynamics [Fig. \ref{fig:flight_experiment_fig_1}]. To mitigate this, a static pitching-up moment offset on the hardware may be compensated for with identification of the coefficient $c_{\tau,0}$ [see Section \ref{sec:aero_model}] from flight data. However, dynamic pitch-rate dependencies and dynamic stall phenomena further play a role for the experienced high-rate maneuvers and can be included in the aerodynamic load model. 

Furthermore, we observed that at initial speeds of 5 m/s and in particular in a case of minimizing changes in forward velocity, the pitch angle must be increased significantly by the controller to prevent stall whilst maintaining flight altitude through the gap. This poses the risk of tail strike with the lower edges of the gap and can limit gap-passage success at flight speeds less than 5 m/s. To provide general safety guarantees against tail collision, a maximum pitch angle constraint during gap passage may be embedded into the optimization and MPC framework. 

In conclusion, we described design, modeling, and control methods to achieve accurate, collision free narrow-gap passage with wing-sweep morphing drones. We proposed and validated an aerodynamic model that enables real-time control of wing-sweep morphing drones in low-Reynolds and post-stall flight regimes. We used this model to perform trajectory optimization under different flight optimization criteria and compared flight behaviors emerging from the trajectories. Finally, we performed flight experiments on a lightweight drone prototype. Nonlinear MPC with adaptive constraint handling capabilities realizes successful gap passage with an average altitude error of 5 cm to the center of the gap, even when the forward flight speed remains unchanged throughout. Wing-sweep morphing actuation hence shows promise for increasing the versatility of winged drones in traversing narrower gaps in cluttered environments, supporting applications in aerial mobility, search and rescue, and indoor flight missions.

\section{Acknowledgments}

% \centering
% \textit{Excluded for double-anonymous review process}

The authors acknowledge Valentin Wüest for helpful insights during control software development and thank Simon Jeger for assistance with the experimental setup and discussions on the lateral controller. This work was partly funded by the Swiss National Science Foundation project "Avian-informed Drones" nr. 10.002.693.

\bibliographystyle{IEEEtran}
\bibliography{IEEEabrv,references}

\end{document}